# Content-Augmented Feature Pyramid Network with Light Linear Spatial Transformers for Object Detection

Yongxiang Gu[1,4], Xiaolin Qin[1,2,4, *], Yuncong Peng[1,4] and Lu Li[3]

[1] Chengdu Institute of Computer Applications, Chinese Academy of Sciences, Chengdu, 610041, China
[2] Nanchang Institute of Technology, Jiangxi Nanchang, 330044, China
[3] Zenseact, Gothenburg, 41756, Sweden
[4] University of Chinese Academy of Sciences, Beijing, 100049, China
*Corresponding Author: Xiaolin Qin. Email: qinxl2001@126.com

**Abstract:** As one of the prevalent components, Feature Pyramid Network (FPN) is widely used in current object detection models for improving multi-scale object detection performance. However, its feature fusion mode is still in a misaligned and local manner, thus limiting the representation power. To address the inherit defects of FPN, a novel architecture termed Content-Augmented Feature Pyramid Network (CA-FPN) is proposed in this paper. Firstly, a Global Content Extraction Module (GCEM) is proposed to extract multi-scale context information. Secondly, lightweight linear spatial Transformer connections are added in the top-down pathway to augment each feature map with multi-scale features, where a linearized approximate self-attention function is designed for reducing model complexity. By means of the self-attention mechanism in Transformer, there is no longer need to align feature maps during feature fusion, thus solving the misaligned defect. By setting the query scope to the entire feature map, the local defect can also be solved. Extensive experiments on COCO and PASCAL VOC datasets demonstrated that our CA-FPN outperforms other FPN-based detectors without bells and whistles and is robust in different settings.

## 1 Introduction

Computer vision has been greatly accelerated by deep learning since 2012, and many important applications like self-driving cars become more and more investigated by industry ever since. Such applications require to solve the challenging task of object detection, which not only to classify an image, but also to accurately predict the 2D/3D boundary of object of interest in the image. For the former, it requires to extract high-level features from an image, and the latter depends on the low-level features. Thanks to Feature Pyramid Network (FPN) [1], which achieves a good tradeoff by fusing features at different levels and has been adopted as a standard building block for constructing object detection models.

To further improve the performance of FPN, we can gain some intuitions from the human vision system. When people are uncertain about an object in an image, they will naturally search for similar objects and combine the background information to make a more comprehensive judgment (see Fig. 1). Therefore, we hypothesized that the capability of association is the key for further improvements of detection performance. Technically, although a series of FPNs [2-6] have been proposed to make continuous improvement, there still exist two main inherent defects in FPN, limiting the above capability.

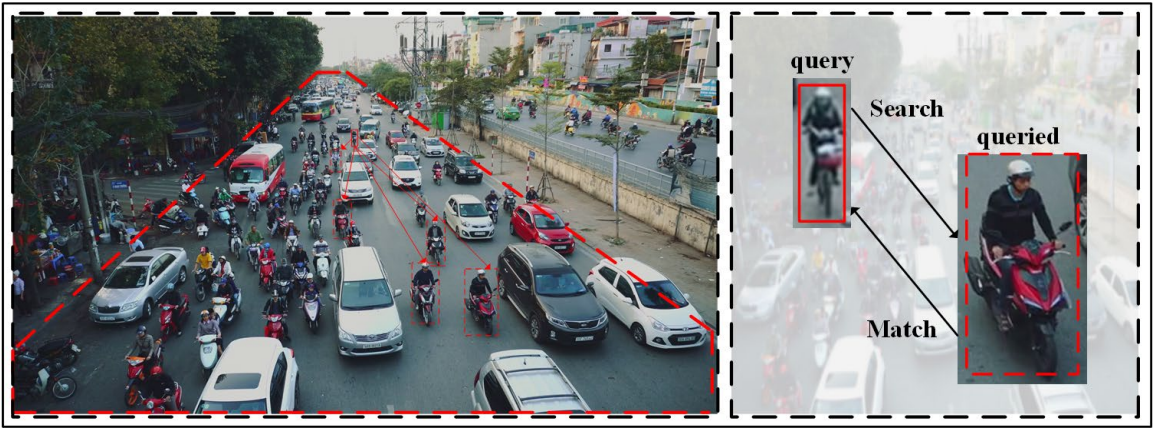

**Figure 1:** Global view of human vision in object detection. To distinguish the indistinct "people" instance in the solid box, other similar and clear "people" instances in the dashed boxes can be used to enhance our judgment. Meanwhile, the background information (the big dotted box on the left) is also crucial, since people are more likely to be on city roads than trees.

1) **The feature fusion mode is misaligned.** When FPN fuses multi-scale features in the top-down pathway, the coarser-resolution feature map needs to be up-sampled by interpolation or learnable convolutions (e.g., transposed convolution). However, the mapping from high-resolution to low-resolution is generally irreversible, leading to the fused feature map with misaligned contexts, in turn, causing deterioration in localization ability, especially on object boundaries [7]. What's worse is that the way of interpolation may bring new issues like alias effect [1], while learnable convolutions introduce non-ignorable computation overhead. In short, the misaligned defect makes FPN difficult to search similar instances and obtain precise background information.

2) **The feature fusion mode is local.** Convolutional Neural Networks (CNNs) are characterized by local connections. Although stacking convolutional layers can increase the receptive field size to some extent, the effective receptive field size increases linearly with the square root of the number [8], determining the local property of each convolutional feature map. On the other hand, FPN simply conducts pixel addition or 1×1 convolution at each corresponding spatial location to fuse multi-scale features. Consequently, the fused feature vector can merely integrate information within a set of nearby locations, determining the local property of each fused feature map. In short, the local defect poses the fundamental challenge for FPN to obtain the global background information and search for similar instances.

In principle, the misaligned defect is caused by down-sampling and up-sampling operations. Therefore, HRNet [9] maintains high-resolution representations to fundamentally eliminate sampling operations but brings new issues of complicated computation and poor real-time performance. To efficiently alleviate the misaligned defect, FaPN [7] proposes a feature alignment module to adjust each sampling location by learning transformation offsets.

On the other hand, aiming at the locality property of convolutions, NLNet [10] proposes non-local operations to capture remote dependencies directly by calculating the similarity function between any two positions, which shares the similarity with self-attention [11]. However, NLNet faces the huge computation and space overhead, and has not solved the local defect of feature fusion. Recently, Transformer-based methods achieve remarkable performance in computer vision [6], by capturing long-range dependencies with self-attention mechanism, and overcome the locality property of convolutions naturally. However, the high model complexity limits its application, thus there are few works to introduce Transformer in FPN and rarely preserve its global query scope at the same time.

To solve the inherit detects of FPN, we note that the way of modeling the pairwise relation between features by self-attention can naturally fit as a solution. Specifically, we firstly introduce deformable convolution networks v2 (DCNv2) [12] and spatial attention module (SAM) [13] into DenseASPP [14] to build a strong global content extraction module (GCEM), which can generate a feature map with multi-scale context information (denoted as $P_G$). Then, we add lightweight linear spatial Transformer connections between $P_G$ and each feature map in the top-down pathway, where the former is set as the queried map and the latter is set as the query map. By means of the self-attention mechanism, there is no longer need to align feature maps during feature fusion, thus solving the misaligned defect. By setting the query scope of the self-attention mechanism to the entire feature map, the local defect can also be solved.

In the global feature interaction of Transformer connections, high-matched features can provide the similar instance features while a large number of low-matched features can provide background context information, largely augmenting each query feature map. Therefore, the proposed CA-FPN is functionally closer to the human vision system. We highlight our principal contributions as follows:

1) To provide multi-scale context information for augmenting the representation power, we propose a Global Content Extraction Module (GCEM) and equip it on top of the FPN.

2) To address the misaligned and local defects of feature fusion mode in FPN, we deeply combine our GCEM with FPN through lightweight linear spatial Transformer connections.

3) A linearized approximate self-attention function is designed in linear Transformer for reducing model complexity.

4) We construct a novel architecture termed Content-Augmented Feature Pyramid Network (CA-FPN) and outperform other FPN-based detectors on the challenging COCO and PASCAL VOC datasets.

## 2 Related Work

In this section, we first discuss FPN, and then focus on scale-aware modules and Transformer, which can be components of FPN. Actually, the proposed CA-FPN is exactly constructed with a scale-aware model and Transformer connections base on the vanilla FPN. Finally, other optimization modules used in this paper are briefly introduced, including deformable convolutions and attention modules.

### 2.1 Feature Pyramid Network

As shown in Fig. 2, the generalized FPN consists of multiple top-down and bottom-up pathways and can equip with an additional scale-aware module. In vanilla FPN [1], a top-down pathway with lateral connections is introduced to fuse multi-scale features. Since both localization information and semantic information can be considered in the cross-scale feature fusion, FPN shows great advances for detecting objects with a wide variety of scales [15].

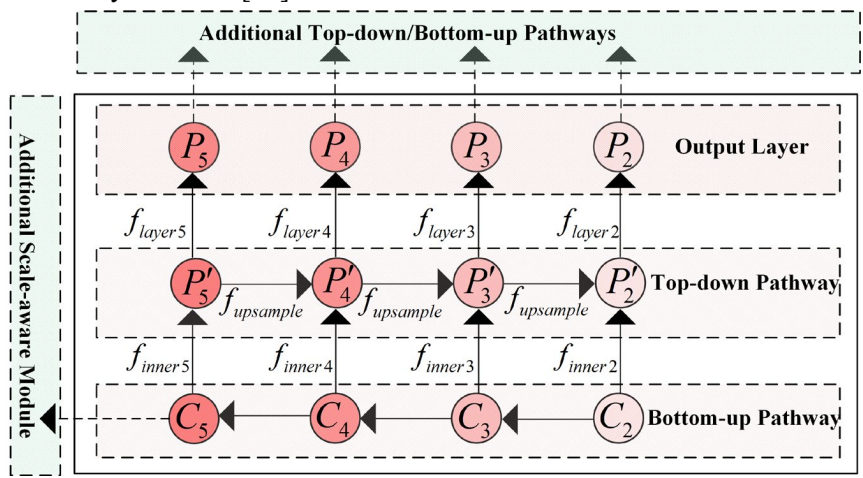

**Figure 2:** Architecture of the generalized FPN.

In recent years, a multitude of FPN structures have been proposed [2-6]. In PANet [2], the extra Bottom-up Path Augmentation (BPA) and the adaptive feature pooling are proposed for boosting information flow and aggregating features. In Libra R-CNN [3], a deep integrated and balanced semantic feature map is utilized to strengthen each feature map in vanilla FPN. In NAS-FPN [4], Neural Architecture

Search (NAS) with reinforcement learning is utilized to automatically design the optimal FPN structure for RetinaNet. Although NAS-FPN achieves satisfying performance, the search process is extremely time-consuming and the information flow is difficult to interpret. In EfficientDet [5], Bi-FPN is proposed by simplifying nodes that contribute little to feature fusion and adding cross-layer connections based on PANet.

Although these FPNs have made continuous improvement in optimizing feature fusion, they mainly focus on adding fusion paths or removing existing weak fusion paths, while the basic operations of aligning feature maps and integrating information within a set of nearby locations are not changed. In this paper, to solve the inherit detects of FPN, we propose CA-FPN by the core idea of adding Transformer connections in the top-down pathway.

*2.2 Scale-aware Modules*

For tasks such as object detection and semantic segmentation, both large-scale (global) and small-scale (local) features need to be integrated. Therefore, a scale-aware module is crucial, which can explicitly optimize multi-scale feature representations. As shown in Fig. 2, scale-aware modules can be equipped on top of the FPN to extract multi-scale features and increase the receptive field.

Spatial Pyramid Pooling (SPP) [16] module is a classic scale-aware module, which extracts multi-scale features through pooling layers with different receptive fields. In order to balance global and local features, DeepLabv2 [17] proposes the Atrous Spatial Pyramid Pooling (ASPP) module for semantic segmentation tasks, which extracts multi-scale features in parallel through a set of atrous (dilation) convolutions with different atrous (dilation) rates. Although ASPP is able to generate rich multi-scale features, DeepLabv3 [14] pointed out that the feature resolution in the scale-axis is not dense enough in practice. Therefore, DeepLabv3 further proposes the Dense Atrous Spatial Pyramid Pooling (DenseASPP) module, which uses dense connections [18] to replace the parallel feature extraction pathways in ASPP, and obtains stronger multi-scale feature representation in a dense way.

In this paper, on account of the significance of context information, we further introduce DCNv2 [12] and SAM [13] into DenseASPP to build a strong global content extraction module (GCEM) and equip it on top of the FPN. Specifically, the feature map with multi-scale context information generated by GCEM will serve as the queried map to augment each feature map in FPN through Transformer connections.

*2.3 Transformer*

Transformer is designed in Natural Language Processing (NLP) for sequence modeling and transduction tasks. It is notable for its use of attention to model long-range dependencies [11]. Recently, Transformer-based methods have achieved remarkable performance in computer vision [6], rising a wave of studies. Transformer is characterized by self-attention mechanism, multi-head attention mechanism, position-wise feed-forward network, layer normalization, and residual connections [19]. Among them, self-attention is the core design of Transformer, which will be detailly introduced in this section.

Given a feature map with resolution of $H \times W$ and feature dimension of $C$, which is the common format in vision domain. We first reshape it to the sequence format and denote it as $x \in R^{HW \times C}$. Formally, the input $x$ is projected by three learnable linear matrixes $W_Q, W_K, W_V$ to corresponding representations $Q$, $K$, and $V$: $Q = xW_Q$, $K = xW_K$ and $V = xW_V$. Following common terminology, $Q$, $K$, and $V$ are referred to as query, key, and value respectively.

Then, a transformer block $T: R^{HW \times C} \rightarrow R^{HW \times C}$ with input $x$ is defined as:

$$T(x) = F(A(Q,K,V) + x) \tag{1}$$

where $F(\cdot)$ is a position-wise feed-forward network with a residual connection; $A(\cdot)$ is the self-attention function that computes the attention matrix.

If $M_i$ is used to present the i-th row of the matrix $M$, then the output $O \in R^{HW \times C}$ of $A(x)$ can be computed as:

$$O = A(Q,K,V) = [O_1,...,O_N]^T \tag{2}$$

$$O_i = \sum_j \frac{S(Q_i, K_j)}{\sum_j S(Q_i, K_j)} V_j \tag{3}$$

When $S(Q,K) = e^{QK^T}$, we can get the standard self-attention in vanilla Transformer, i.e., the dot-product attention with softmax normalization. In this case, the computation and space complexity to compute each row the output $O_i$ is $O(HWC)$. Thus, the overall computation and space complexity of the output $O$ grows quadratically with the sequence length [20], i.e., the spatial resolution of the feature map in vision domain. The detailed computation and space complexity of vanilla Transformer (VT) are shown as follows (we omit softmax computation in determining the complexity):

$$\Omega_C(VT) = 4HWC^2 + 2H^2W^2C \tag{4}$$

$$\Omega_S(VT) = H^2W^2 \tag{5}$$

Although Transformer has showed remarkable performance in NLP. It poses a great challenge to adopt Transformer from language to vision due to the data difference in two domains, i.e., high resolution in pixels in images compared to words in text [6]. It means that standard self-attention operations will bring huge computation and space overhead for vision models.

To reduce model complexity, CCNet [21] uses a cris-cross attention module to capture context information on its cris-cross path. ISSA [22] borrows the idea of interlace mechanism, decomposing the dense similarity matrix into the product of two sparse similarity matrixes. Since the principal computation overhead is brought by the non-linear similarity function, Katharopoulos et al. [23] proposes directly replacing the softmax with a linear operation, which achieves the linear complexity but meets the generalization problem on many NLP downstream tasks. Swin Transformer [6] reduces model complexity by limiting the field of view to the fixed and using shift windows to obtain a dynamic and larger receptive field, and achieves the state-of-the-art performance on COCO. As far as we know, FPT [24] makes an early attempt to introduce Transformer in FPN designs but constrains the query scope to the locality.

In this paper, we approximate the standard similarity function by replacing the softmax with a linear approximate operation, largely reducing model complexity. Therefore, we add linear Transformer connections in the top-down pathway to augment each feature map efficiently and preserve the global query scope at the same time.

*2.4 Optimization Modules*

Currently, DCNs and attention modules are widely used in object detection models, acting as the optimization modules for further performance improvement.

**DCNs.** Traditional CNNs extract features within a fixed square receptive field, limiting the ability of geometric transformations modeling. In view of this problem, Deformable Convolutional Networks (DCNs) [8] augment the spatial sampling locations with additional offsets and learn the offsets from the target tasks. Furthermore, DCNv2 [12] introduces a modulation mechanism in DCNs to expand the scope of deformation modeling.

**Attention Modules.** In the past decade, attention modules play an increasingly improtant role in computer vision, which can implicitly and adaptively predict the potential key features through statistical information [25]. SENet [26] pioneered channel attention, which extracts global spatial information by global average pooling and then captures channel-wise relationships by the excitation module. Inspired by SENet, CBAM [13] expands attention mechanism from channel dimension to space dimension and proposes the spatial attention module (SAM). In addition, CBAM further leverages discriminant information by introducing max pooling.

## 3 Method

Motivated by the human vision system, we decide to augment FPN with the capability of association. However, there exist misaligned and local defects in FPN, limiting the feature representation power. To address these, we note that the way of modeling the pairwise relation between features by self-attention can naturally fit as a solution. Therefore, we construct a novel architecture termed CA-FPN. We elaborate on this network in this section.

### 3.1 Overview

Built upon the vanilla FPN [1], our CA-FPN has two novel components: 1) Global Content Extraction Module (GCEM) and 2) Linear Transformer (LT). The former is designed for providing multi-scale context information for the query maps, while the latter is adapted to address the inherent defects of FPN.

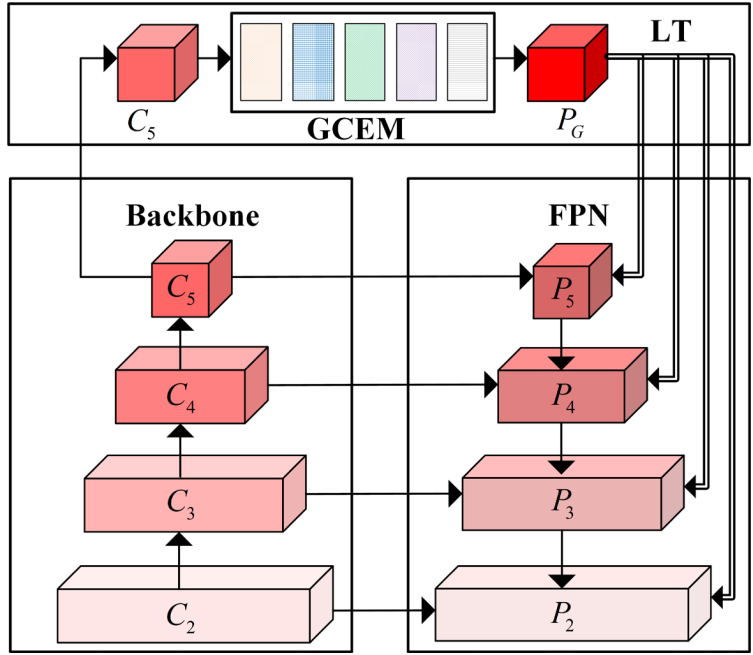

**Figure 3:** The overall architecture of CA-FPN

As shown in Fig. 3, GCEM is deeply combined with the vanilla FPN through light linear spatial transformers. Specifically, the output of the i-th stage of the bottom-up network is defined as $C_i$, which has stride of $2^i$ pixels with respect to the input image; the feature map generated by GCEM is defined as $P_G$, whose channel number is 256. In vanilla FPN, a 1×1 convolutional layer is attached on $C_i$ to generate the temporary feature map $P'_i$ (as shown in Fig. 2) with the fixed channel dimension of 256. We take $P'_i$ as the query map and take $P_G$ as the queried map. It means that $Q$ is linearly projected from $P'_i$, while $K$ and $V$ are linearly projected from $P_G$. Here, $P'_i$ and $P_G$ are reshaped to the corresponding sequence format to meet the operational requirements. In this paper, we modify the transformer block $T: R^{HW \times C} \to R^{HW \times C}$ to the more generalized definition with the query input $P'_i$ and the queried input $P_G$, as follows:

$$T(P'_i, P_G) = F(A(Q, K, V) + P'_i) \tag{6}$$

$$Q = P'_i W_Q, K = P_G W_K, V = P_G W_V \tag{7}$$

As shown above, to address the inherit defects of FPN, we add spatial Transformer connection between $P'_i$ and $P_G$ to generate the targeted content-augmented information for $P'_i$. As a result, objects with different scales can learn targeted context information. By means of the self-attention mechanism, there is no longer need to align feature maps during feature fusion, thus solving the misaligned defect. By setting the query scope of the self-attention mechanism to the entire feature map, the local defect can also be solved.

### 3.2 Global Content Extraction Module

Context information is crucial for making synthetic judgments since it can provide additional relevant information. It becomes more important in this paper since multi-scale context information is used to provide targeted information for each query. To build a global content extraction module (GCEM), we introduce DCNv2 [12] and SAM [13] into DenseASPP [14] to augment its ability of extracting features.

**DCNv2.** Although DenseASPP has shown great ability in extracting multi-scale features, it seems inflexible to extract features within a number of fixed and predefined receptive fields. To address this, we introduce DCNv2 to replace each atrous convolution in DenseASPP. Note that the modulation mechanism gives each DCNv2 the ability to be a generalized atrous convolution with the dynamic atrous rate and the rate value is purely learned from the context according to the task. Besides, the additional sampling offset enhances the ability to extract boundary information. Therefore, compared to the counterpart with atrous convolutions, a scale-aware module stacked with a series of DCNv2 modules can extract richer multi-scale features in theory.

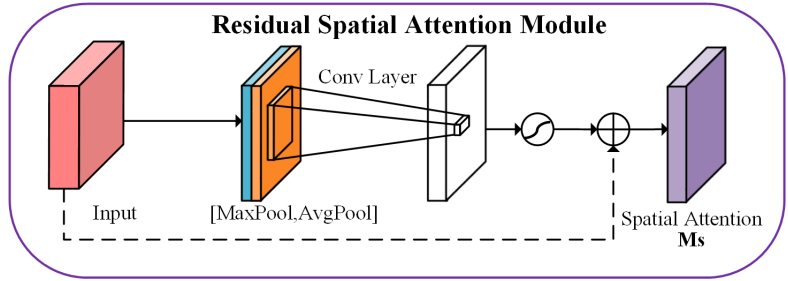

**Figure 4:** Architecture of residual SAM

**SAM.** In the DenseASPP-like structure, each DCNv2 can capture the boundary information at the specific scale. Directly fusing plenty of features with a 1×1 convolutional layer in DenseASPP seems rough and may rise localization confusion. To reweight the importance of spatial locations, we introduce SAM to refine the features. As shown in Fig. 4, a residual connection is added to further facilitate the flow of information at a low cost. Since GCEM is equipped on top of the FPN, channels of the deep feature map can clearly respond to category information, therefore, we no longer introduce channel attention modules.

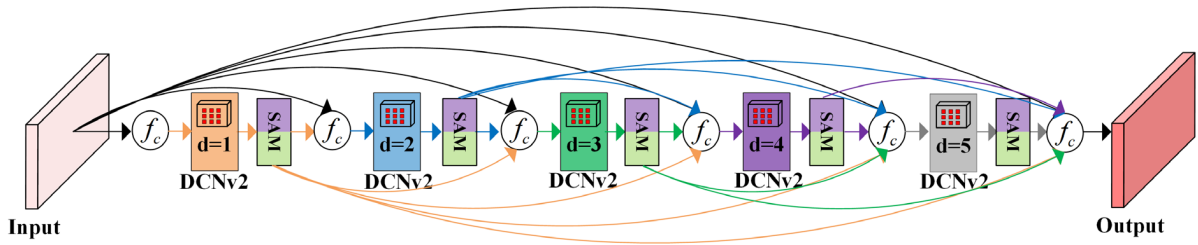

**Figure 5:** Architecture of GCEM

As shown in Fig. 5, our GCEM consists of a basic module containing the compressing function $f_c$, DCNv2, and a residual SAM, which is stacked five times with dense connections in total. Specifically, $f_c$ first concatenates the input features in channel dimension, and uses a 1×1 convolutional layer to compress the channel dimension to 512 except that the final one compresses the channel dimension to 256. Then, DCNv2 adjusts the channel dimension from 512 to 256 and SAM refines the features without changing the channel dimension. Following the design principle in DenseASPP, we control the model size by channel reduction and reduce the channel dimension to the half each time. Since the channel dimension of each query feature map is 256, we design the channel reduction strategy by retrodiction. Following the past experience in DenseASPP, i.e., gradually enlarging the atrous rates, we first implement DCNv2 based on the atrous convolution instead of the traditional convolution, and then set the atrous rates of 5 basic modules as 1-5 to get a default initialization. Specifically, we input $C_5$ to GCEM and use the output as the queried feature map in Transformer connections.

## 3.3 Linear Transformer

Although the idea of adding Transformer connections in the top-down pathway to augment each feature map with multi-scale features is promising, we have to face huge computation and space overhead brought by vanilla Transformer.

Look back to Eq. 3, any similarity function can be selected to calculate the attention matrix. To achieve the linear complexity of the sequence length, a decomposable similarity function can be adopted, as follows:

$$S(Q_i, K_j) = \varphi(Q_i)\varphi(K_j)^T \tag{8}$$

where $\varphi(\cdot)$ is the kernel function to project $Q$ and $K$ to the hidden representations. Then, Eq. 3 can be formed as:

$$O_i = \frac{\sum_{j=1}^{N}\left(\varphi(Q_i)\varphi(K_j)^T\right)V_j}{\sum_{j=1}^{N}\left(\varphi(Q_i)\varphi(K_j)^T\right)} \tag{9}$$

For successive matrix multiplication of $Q$, $K^T$, $V$, we can simplify operations by using the commutative law of matrix multiplication:

$$\left(\varphi(Q)\varphi(K)^T\right)V = \varphi(Q)\left(\varphi(K)^T V\right) \tag{10}$$

Instead of computing the attention matrix $QK^T \in R^{HW \times HW}$, we can first compute $\varphi(K)^T V \in R^{C \times C}$ and then multiply with $\varphi(Q) \in R^{HW \times C}$. As a result, computation and space complexity can be reduced to $O(C^2 HW)$ and $O(C^2)$. Since $C \ll HW$, overall complexity can be largely optimized. [20] The main difficulty of this idea is to select the appropriate decomposable similarity function, which owns practical interpretability and is easy to implement. Along this direction, previous works mainly focus on designing the kernel function [23] separately, while ignore the consideration of taking the decomposable similarity function as a whole.

**Approximation.** To design the appropriate decomposable similarity function, we look back to the standard similarity function and conduct Taylor expansions on it, as follows:

$$S(Q_i, K_j) = e^{Q_i K_j^T} = \sum_{n=0}^{\infty} \frac{\left(Q_i K_j^T\right)^n}{n!} \quad \forall Q_i K_j^T \tag{11}$$

Note that the first Taylor order expansion of the above equation meets the decomposable property naturally. Therefore, we decide to take it as the similarity function, i.e., $e^{Q_i K_j^T} \approx 1 + Q_i K_j^T$. Furthermore, to meet the non-negative requirement of the similarity function, we normalize $Q_i$ and $K_j$ by dividing them with the corresponding L2 norms and denote them as $\bar{Q}_i$ and $\bar{K}_j$, respectively.

$$\bar{Q}_i = \frac{Q_i}{\|Q_i\|_2}, \bar{K}_j = \frac{K_j}{\|K_j\|_2} \tag{12}$$

By substituting the normalized $\bar{Q}_i$ and $\bar{K}_j$ to the first Taylor order expansion, we get the approximate similarity function $S(Q_i, K_j) = 1 + \bar{Q}_i \bar{K}_j^T$. Correspondingly, the Eq. 3 can be rewritten as follows:

$$O_i = \frac{\sum_j \left(V_j + \bar{Q}_i\left(\bar{K}_j^T V_j\right)\right)}{HW + \bar{Q}_i \sum_j \bar{K}_j^T} \tag{13}$$

To reduce complexity of Transformer, we directly approximate the self-attention function by replacing the softmax normalization with a linear approximate operation, i.e., its first Taylor order expansion. It makes the proposed similarity function more interpretable. With the proposed approximate self-attention function, we further build the linear Transformer (LT).

The detailed computation and space complexity of LT are shown as follows

$$\Omega_C(LT)=4HWC^2+2HWC^2 \tag{14}$$

$$\Omega_S(LT)=C^2 \tag{15}$$

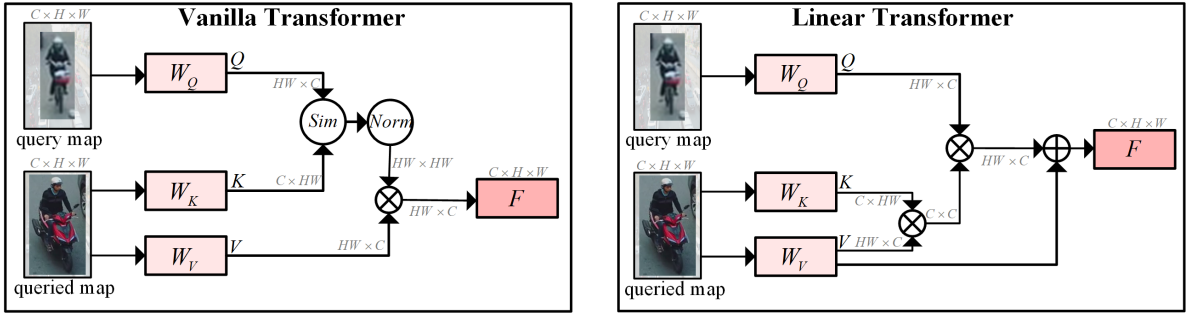

**Figure 6:** Structures of vanilla Transformer and linear Transformer

As shown in Fig. 6, the proposed Transformer is easy to implement with direct and simple matrix operations of matrix multiplication and matrix addition. It makes the proposed method of a broad application prospect under the premise of controlling approximation loss within an acceptable range.

*3.4 Summary*

The details of CA-FPN are introduced in this section. Specifically, DCNv2 and SAM are introduced into DenseASPP to build a strong global content extraction module (GCEM). In addition, a linearized attention function is designed in linear Transformer (LT) by the idea of approximation. Finally, CA-FPN is constructed by deeply combining the GCEM with the vanilla FPN through LT.

**4 Experiments**

In this section, we will introduce our experiments in detail, including datasets, experiment settings, evaluation metrics, result analysis, ablation studies and effectiveness studies.

*4.1 Dataset*

Experiments are conducted on COCO 2017 [27] and PASCAL VOC 2007, 2012 [28] dataset. Specifically, the challenging COCO dataset has 80 classes, containing 118k training images (trainval35k) and 5k validation images (minival), while PASCAL VOC has 20 classes. Following the normal practice in the literature, when training on PASCAL VOC, the models are trained on the union of VOC2007 and VOC2012 trainval set and tested on VOC 2007 test set. Since we focus on object detection in this paper, we train the independent object detector without using the mask information.

**Table 1:** Statistics of COCO and PASCAL VOC datasets.

| Dataset | Number of classes | Training Images | Testing Images |
|---|---|---|---|
| COCO 2017 | 80 | 118287 | 5000 |
| PASCAL VOC 2007, 2012 | 20 | 16551 | 4952 |

*4.2 Experiment Setting and Evaluation Metric*

We evaluate CA-FPN on COCO and PASCAL VOC datasets with different detectors (Faster R-CNN [29], RetinaNet [30], and Cascade R-CNN [31]), which are the representations of the two-stage, one-stage, and multi-stage detectors. The classical networks ResNet-50 and ResNet-101 [19] pre-trained on ImageNet [32] are adopted as backbones for comparative experiments.

Experiments are implemented based on detectron2 [33]. The input images are resized such that the shorter side has 800 pixels and the longest side has 1333 pixels. We train models with ResNet-50 on 1 NVIDIA 2080ti GPU and train other models on 1 NVIDIA 3090 GPU. Restricted by hardware, we set batch-size as 2 instead of 16 and follow the linear scaling rule for adjusting the learning rate [34]. In 1× training period (~12 epochs), there are 720k iterations in total. The learning rate is divided by 10 at 480k and 640k. In 3× training period (~37 epochs), we triple the total iteration number to 2160k and the learning rate is divided by 10 at 1680k and 2000k. In all training periods, we take the linear warm-up strategy in the first 8k iterations. Specifically, when training on PASCAL VOC, the total epochs are set as 17.4.

For evaluation, we adopt the standard evaluation metrics, i.e., AP, $AP_{50}$, $AP_{75}$, $AP_S$, $AP_M$ and $AP_L$. The last three measure performance with respect to objects of small, medium, and large sizes. Following the practice in the literature [35], FLOPs are reported to measure the computation overhead and training memory is additionally reported in the effectiveness studies on Transformers for further analysis.

*4.3 Result Analysis*

**Table 2:** Comparisons using Faster R-CNN in 1× training period on COCO 2017 mini-val.

| Faster R-CNN | Backbone | Params | FLOPs | AP | $AP_{50}$ | $AP_{75}$ | $AP_S$ | $AP_M$ | $AP_L$ |
|---|---|---|---|---|---|---|---|---|---|
| C4 [29] | ResNet-50 | 33.8M | 547.3G | 35.7 | 56.1 | 38.0 | 19.2 | 40.9 | 48.7 |
| FPN* [1] | | 41.7M | 180.0G | 38.0 | 58.8 | 41.4 | 22.5 | 41.5 | 49.4 |
| PANet [2] | | 46.4M | 201.9G | 38.7 | 60.4 | 41.7 | 22.6 | 42.6 | 50.3 |
| FPN +Dense ASPP* [14] | | 54.6M | 191.3G | 39.4 | 61.0 | 42.9 | **23.5** | 42.9 | 51.3 |
| Libra R-CNN [3] | | 42.0M | - | 38.7 | 59.9 | 42.0 | 22.5 | 41.1 | 48.7 |
| CA-FPN (ours) | | 55.9M | 197.4G | **40.5** | **61.7** | **44.1** | 23.1 | **44.1** | **53.9** |
| C4* [29] | ResNet-101 | 52.7M | 729.3G | 39.2 | 59.2 | 42.0 | 20.8 | 44.3 | 54.2 |
| FPN* [1] | | 60.6M | 247.1G | 40.3 | 61.5 | 43.9 | 24.1 | 44.2 | 51.7 |
| PANet [2] | | 65.3M | 268.9G | 40.5 | 62.0 | 43.8 | 23.0 | 44.8 | 53.2 |
| FPN +Dense ASPP* [14] | | 73.5M | 258.4G | 41.3 | 62.9 | 45.0 | 24.6 | 45.3 | 53.0 |
| Libra R-CNN [3] | | 60.9M | - | 40.3 | 61.3 | 43.9 | 22.9 | 43.1 | 51.0 |
| CA-FPN (ours) | | 74.8M | 264.5G | **41.9** | **63.1** | **45.7** | **24.7** | **45.4** | **55.7** |

*Note:* * means our re-implemented results.

As shown in Tab. 2, our method outperforms FPN by a large margin. By replacing FPN with CA-FPN, Faster R-CNN gains 2.5 and 1.6 points improvement on ResNet-50 and ResNet-101. What's more, our CA-FPN built upon ResNet-50 even achieves comparable performance with FPN built upon ResNet-101. Comparing with other advanced methods, CA-FPN is still advantageous, especially for big objects. The overall performance improvement of our method is in line with expectations. Although objects of all scales can gain background augmentation and aggregate information from similar objects, large objects with more spatial pixels interact more with the whole feature map.

**Table 3:** Comparisons using Faster R-CNN on PASCAL VOC 2007 test set.

| Faster R-CNN | Backbone | Params | FLOPs | AP | $AP_{50}$ | $AP_{75}$ | $AP_S$ | $AP_M$ | $AP_L$ |
|---|---|---|---|---|---|---|---|---|---|
| C4* [29] | ResNet-50 | 33.2M | 414.3G | 47.0 | 76.7 | 50.5 | 15.7 | 36.9 | 54.5 |
| FPN* [1] | | 41.4M | 179.7G | 49.4 | 79.2 | 53.8 | 19.0 | 39.0 | 56.4 |
| CA-FPN (ours) | | 54.3M | 197.1G | **52.2** | **79.4** | **57.6** | **20.0** | **40.1** | **60.1** |
| C4* [29] | ResNet-101 | 52.1M | 702.9G | 51.0 | 78.3 | 55.7 | 16.5 | 39.5 | 59.0 |
| FPN* [1] | | 60.3M | 246.8G | 51.6 | 79.4 | 57.0 | 19.4 | 41.3 | 58.9 |
| CA-FPN (ours) | | 73.2M | 264.2G | **54.3** | **81.0** | **60.4** | **21.1** | **42.3** | **62.1** |

*Note:* * means our re-implemented results.

Tab. 3 shows the detection performance on PASCAL VOC 2007 test benchmark. By replacing FPN with CA-FPN, Faster R-CNN achieves 52.2 and 54.3 AP on ResNet-50 and ResNet-101, which is 1.9 and 2.7 points higher than the baseline. On both COCO and Pascal VOC datasets, the accuracy steadily improves when leveraging CA-FPN, especially for big objects.

**Table 4:** Comparisons using Faster R-CNN in 3× training period on COCO 2017 mini-val set.

| Faster R-CNN | Backbone | Params | FLOPs | AP | $AP_{50}$ | $AP_{75}$ | $AP_S$ | $AP_M$ | $AP_L$ |
|---|---|---|---|---|---|---|---|---|---|
| C4 [29] | ResNet-50 | 33.8M | 547.3G | 38.4 | 58.7 | 41.3 | 20.7 | 42.7 | 53.1 |
| FPN* [1] |  | 41.7M | 180.0G | 40.0 | 60.6 | 43.6 | 23.9 | 43.3 | 52.2 |
| CA-FPN (ours) |  | 55.9M | 197.4G | **42.0** | **63.3** | **45.6** | **25.5** | **45.2** | **55.4** |
| C4 [29] | ResNet-101 | 52.7M | 729.3G | 41.1 | 61.4 | 44.0 | 22.2 | 45.5 | 55.9 |
| FPN* [1] |  | 60.6M | 247.1G | 42.0 | 62.5 | 45.9 | 25.2 | 45.6 | 54.6 |
| CA-FPN (ours) |  | 74.8M | 264.5G | **43.1** | **64.0** | **47.0** | **26.2** | **46.8** | **56.5** |

*Note:* * means our re-implemented results.

As shown in Tab. 4, the improvement brought by CA-FPN is consistent as the training period lengthened. By replacing FPN with CA-FPN in 3× training, Faster R-CNN gains 2.0 and 1.1 points improvement on ResNet-50 and ResNet-101, respectively. Even for the tough small object detection, CA-FPN can still bring consistent improvement. In Faster R-CNN, CA-FPN built upon ResNet-50 yields 0.6 and 1.6 APs improvement in 1× and 3× training compared to the vanilla FPN. While enlarging the backbone to ResNet-101, CA-FPN yields 0.6 and 1.0 APs improvement in 1× and 3× training.

From Tab. 2-4, an interesting phenomenon can be found that Faster R-CNN built on ResNet-101 with C4 head can perform well on big objects, even surpassing advanced methods. In other FPN-based detectors, big objects generally cannot obtain specific background information, since high-level semantic feature map also contributes to detection of non-corresponding scales. By contrast, we can augment each feature map in FPN with the global content extraction module by linear Transformer connections. As a result, objects at different scales can additionally learn targeted context information.

**Table 5:** Comparisons using RetinaNet and Cascade R-CNN built upon ResNet-50.

| Method | Dataset | + Our modules | Schedule | Params | FLOPs | AP | $AP_{50}$ | $AP_{75}$ | $AP_S$ | $AP_M$ | $AP_L$ |
|---|---|---|---|---|---|---|---|---|---|---|---|
| RetinaNet [30] | PASCAL VOC |  | 17.4e | 37.9M | 206.8G | 51.2 | 77.5 | 55.1 | 18.4 | 38.8 | 59.0 |
|  |  | ✓ |  | 52.1M | 220.2G | **51.9** | **78.2** | **56.0** | **20.5** | **39.8** | **59.3** |
|  | COCO |  | 1× | 37.9M | 206.8G | 37.4 | 56.5 | 40.0 | 21.5 | 41.5 | 47.7 |
|  |  | ✓ |  | 52.1M | 220.3G | **38.7** | **58.8** | **41.2** | **23.9** | **43.5** | **47.8** |
| Cascade R-CNN [31] | PASCAL VOC |  | 17.4e | 69.2M | 207.5G | 54.4 | 78.3 | 59.0 | 20.4 | 42.2 | 62.7 |
|  |  | ✓ |  | 83.4M | 224.9G | **56.0** | **79.4** | **61.0** | **21.4** | **43.5** | **64.2** |
|  | COCO |  | 1× | 69.3M | 207.7G | 41.5 | 59.4 | 45.1 | 24.0 | 44.8 | 54.6 |
|  |  | ✓ |  | 83.5M | 225.0G | **43.0** | **61.4** | **46.5** | **24.8** | **46.7** | **56.7** |
| Cascade R-CNN [31] | COCO |  | 3× | 69.3M | 207.7G | 43.3 | 61.3 | 47.1 | 25.9 | 46.7 | 56.5 |
|  |  | ✓ |  | 83.5M | 225.0G | **44.8** | **63.5** | **48.4** | **28.0** | **48.4** | **58.8** |

Since our CA-FPN can be plugged into existing FPN-based models, it is easy to plug CA-FPN into RetinaNet and Cascade R-CNN. The experiments are conducted on ResNet-50. In Tab. 5, it can be discovered that AP can obtain a significant improvement, which demonstrates that our CA-FPN can provide multi-scale context information for augmenting the representation power. Different from the case of Faster R-CNN, RetinaNet equipped with CA-FPN mainly improves the detection performance of small and medium objects. We speculate that this phenomenon is due to the differences in neck structure. RetinaNet removed the $P_2$ in the neck, thus the semantic gap between $P_G$ and each query feature map is reduced. As a result, context information about small and medium objects can be learned well by linear Transformer connections.

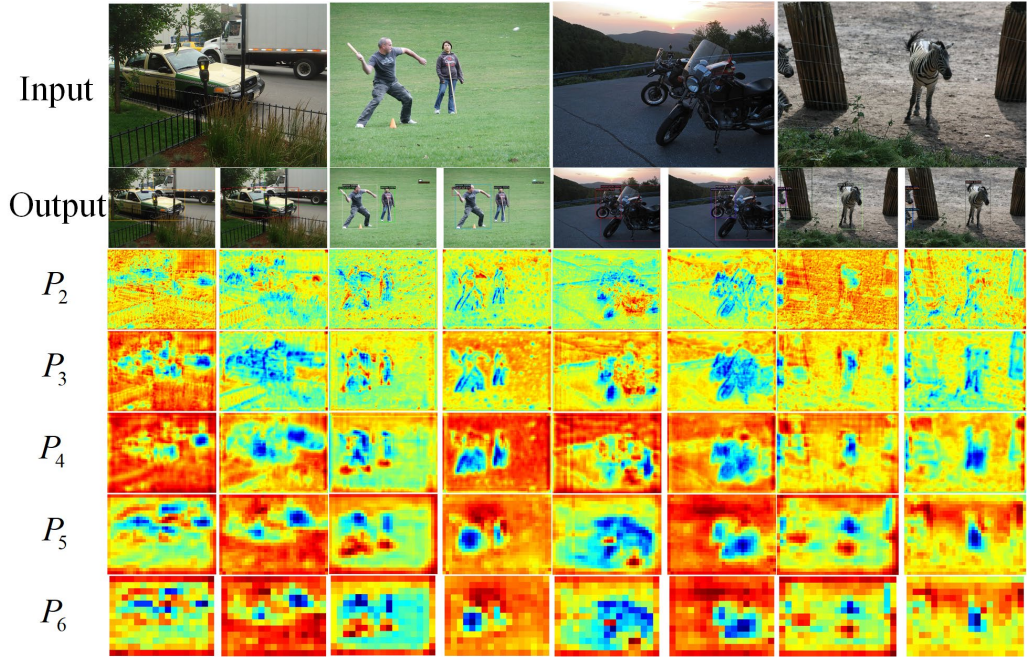

**Figure 7:** Qualitative results of object detection. Both models are built upon ResNet-50 in 1× training period. From top to bottom, there are input images, detection results, and activation maps of $P_2$-$P_6$. The left and right columns under each large input image are the results of vanilla FPN and CA-FPN, respectively.

To vividly demonstrate the feature representation power of CA-FPN, we randomly select images from COCO test set and feed them into detectors. Then, the feature at the 0th channel is visualized on each output of CA-FPN. As shown in Fig. 7, the activated parts of the feature map are highlighted in blue. Take the third image for example, vanilla FPN only mainly activates the wheels, which own the most discriminant information. Differently, benefitting from the global awareness of GCEM and targeted feature interaction capabilities of LT, non-salient relevant parts can also be activated in CA-FPN.

### 4.4 Ablation Studies

In this subsection, we first investigate performance with different ways to build GCEM. Then, we gradually plug GCEM and LT into the vanilla FPN and report the quantitative results. Experiments are conducted on Faster R-CNN using ResNet-50 in 1× training period.

**Ablation Studies on GCEM.** To build a strong scale-aware module, we consider two aspects, i.e., replacing atrous convolutions with DCNv2 modules in DenseASPP and adding the attention module after each DCNv2 module. Channel Attention Module (CAM) and Spatial Attention Module (SAM) proposed in CBAM are chosen as the alternatives owing to their widespread use. To find which type of the attention module is the most suitable, three different ways of arranging attention modules are compared: only the CAM, only the SAM, and the order combination of CAM and SAM (commonly denoted as CBAM). Note that in order to avoid interference from other modules, the proposed LT module is removed from the whole framework. In this case, we conduct point-wise addition between $P_G$ and $P'_5$.

**Table 6:** Ablation studies on GCEM on COCO 2017 mini-val.

| Faster R-CNN | +DCNv2 | +CAM | +SAM | Params | FLOPs | AP | $AP_{50}$ | $AP_{75}$ | $AP_S$ | $AP_M$ | $AP_L$ |
|---|---|---|---|---|---|---|---|---|---|---|---|
| | | | | 54.6M | 191.3G | 39.4 | 61.0 | 42.9 | **23.5** | 42.9 | 42.9 |
| | ✓ | | | 55.1M | 191.9G | 40.0 | 61.1 | 43.5 | 23.3 | 43.2 | **53.5** |
| FPN+DenseASPP [14] | ✓ | ✓ | | 55.2M | 191.9G | 40.1 | 61.2 | **43.6** | 22.8 | 43.2 | **53.5** |
| | ✓ | | ✓ | 55.2M | 191.9G | **40.2** | **61.3** | 43.5 | 22.9 | **43.6** | 53.4 |
| | ✓ | ✓ | ✓ | 55.2M | 191.9G | **40.2** | 60.9 | 43.5 | 23.3 | 43.4 | 53.4 |

Corresponding results are shown in Tab. 6. With DCNv2, the improved method yields 0.6 AP and 10.6 $AP_L$ gains, which indicates that dynamic global content information can greatly improve the detection performance, especially for large objects. Therefore, DCNv2 is chosen as an essential component. Besides, in the last two lines, they clearly show that the improvement brought by SAM is comparable to CBAM. We argue that the reason lies in that, channels of the deep feature map can clearly respond to category information. Therefore, CAM modules contribute little performance gains by adjusting channel weights. Based on the above analysis, we utilize only the SAM, and further combine it with DCNv2 in DenseASPP to build the GCEM.

**Ablation Studies on GCEM and LT.** We breakdown the individual impacts of the two components introduced in CA-FPN by gradually plug them into the vanilla FPN. When only using LT, Transformer connections are added between $P'_5$ and all feature maps in FPN ($P'_2 - P'_5$).

Table 7: Ablation studies on GCEM and LT on MS-COCO 2017 mini-val.

| Faster R-CNN | +GCEM | +LT | Params | FLOPs | AP | $AP_{50}$ | $AP_{75}$ | $AP_S$ | $AP_M$ | $AP_L$ |
|---|---|---|---|---|---|---|---|---|---|---|
| FPN |  |  | 41.7M | 180.0G | 38.0 | 58.8 | 41.4 | 22.5 | 41.5 | 49.4 |
|  | ✓ |  | 55.2M | 191.9G | 40.2 | 61.3 | 43.5 | 22.9 | 43.6 | 53.4 |
|  |  | ✓ | 42.5M | 185.6G | 38.5 | 60.0 | 41.7 | **23.3** | 42.1 | 49.7 |
|  | ✓ | ✓ | 55.9M | 197.4G | **40.5** | **61.7** | **44.1** | 23.1 | **44.1** | **53.9** |

As shown in Tab.7, the model incorporated with both GCEM and LT achieves the best performance in AP, $AP_{50}$, and $AP_{75}$. It should be noted that our GCEM yields 2.2 AP improvement, which shows that the background information is crucial for augmenting the representation power. Moreover, objects with different scales can learn targeted context information with LT. However, it can be found that $AP_S$ drops a little when plugging GCEM into FPN equipped with LT-Taylor. We speculate that this phenomenon is caused by the semantic gap between the query map and the queried map. Since $P_G$ is refined by a multi-layer network, i.e., GCEM, it is difficult for $P_G$ to directly interact with the bottom features through the simple linear transformation in LT-Taylor.

### 4.5 Effectiveness Studies on Transformers

In particular, we investigate the effectiveness of Transformers in two aspects. First, we compare the proposed approximate version with the Vanilla Transformer (VT) which utilizes the dot-product attention with softmax normalization. Further, we present a comparison with another linear Transformer method presented in NLP [20], which adopts $\varphi(\cdot) = ELU(\cdot) + 1$ activation function as the kernel function in Eq. 8. Specifically, we abbreviate the former solution and the proposed approximate version as LT-ELU and LT-Taylor, respectively. It should be noted that LT-ELU utilizes the specific kernel function to decompose the similarity function, while LT-Taylor directly approximates the standard similarity function by replacing the softmax normalization with a linear approximate operation. Finally, we evaluate methods of different Transformer types without GCEM.

Table 8: Effectiveness Studies on Transformers on COCO 2017 mini-val.

| Faster R-CNN | Transformer Type | Params | FLOPs | Train Mem | AP | $AP_{50}$ | $AP_{75}$ | $AP_S$ | $AP_M$ | $AP_L$ |
|---|---|---|---|---|---|---|---|---|---|---|
| FPN+GCEM | w/o Transformer | 55.2M | 191.9G | 5915M | 40.2 | 61.3 | 43.5 | 22.9 | 43.6 | 53.4 |
|  | VT | 55.9M | 231.7G | 7835M | 40.3 | 61.6 | 43.6 | **24.1** | 43.7 | 53.3 |
|  | LT-ELU | 55.9M | 197.4G | 7257M | 40.4 | **61.7** | 43.9 | 23.2 | 43.8 | 53.6 |
|  | LT-Taylor (ours) | 55.9M | 197.4G | 7257M | **40.5** | **61.7** | **44.1** | 23.1 | **44.1** | **53.9** |
| FPN | w/o Transformer | 41.7M | 180.0G | 2843M | 38.0 | 58.8 | 41.4 | 22.5 | 41.5 | 49.4 |
|  | VT | 42.5M | 219.8G | 4716M | **38.9** | **60.6** | **42.0** | **24.0** | **42.3** | 49.6 |
|  | LT-ELU | 42.5M | 185.6G | 4122M | 38.5 | 60.2 | 41.5 | 23.4 | 42.0 | **49.8** |
|  | LT-Taylor (ours) | 42.5M | 185.6G | 4122M | 38.5 | 60.0 | 41.7 | 23.3 | 42.1 | 49.7 |

Corresponding results are shown in Tab. 8. The column of AP clearly shows that incorporating Transformer connections in the top-down pathway of FPN can improve its overall performance. Meanwhile, different similarity function in Transformer shows performance gains in different aspect. Specifically, VT significantly boosts performance in detecting small objects while LT-Taylor works mainly on improving performance in detecting big objects and also brings a bit of gains in detecting small objects. With FPN, linear Transformer methods bring a bit of AP loss versus VT while largely reducing both computation complexity and space complexity. With FPN equipped with GCEM, linear Transformer methods can even surpass the vanilla Transformer in terms of AP and LT-Taylor achieves the best performance in all metrics except $AP_S$.

*4.5 Summary*

In this section, plenty of experiments are conducted to demonstrate the outstanding performance of our CA-FPN. Without bells and whistles, CA-FPN can outperform other competitive FPN-based models and is robust in various settings (backbone, detection head and training period). Through the evaluation on the COCO and PASCAL VOC datasets, we proved that CA-FPN can bring significant improvement in detecting objects at different scales. Through the ablation studies and effectiveness studies, we prove the efficiency and effectiveness of each proposed module with quantitative results.

## 5 Conclusion and Outlook

In this paper, to augment FPN with the capability of association and address the inherent defects in FPN, we construct a novel architecture termed CA-FPN. It is equipped with a global content extraction module and light linear spatial Transformer connections. The former allows to extract multi-scale context information and the latter can deeply combine the global content extraction module with the vanilla FPN using the linearized attention function, which is designed to reduce model complexity. Benefit from the low complexity, we can augment each feature map in FPN with the global content extraction module. As a result, objects with different scales can learn targeted context information. Moreover, our CA-FPN can be readily plugged into existing FPN-based models. Extensive experiments on the challenging COCO and PASCAL VOC dataset demonstrated that our CA-FPN significantly outperforms competitive FPN-based detectors without bells and whistles.

Despite the overall object detection performance of our CA-FPN is well and the computation and space efficiency are especially improved, the overall performance can further be improved with hyperparameter tuning. The subsequent research will focus on how to improve the performance of multi-scale detection in a balanced and all-sided way by better augmenting instance features with context information.

**Funding Statement:** This work was partially supported by National Academy of Science Alliance Collaborative Program (Chengdu Branch of Chinese Academy of Sciences - Chongqing Academy of Science and Technology), National Natural Science Foundation of China (No. 61402537), Sichuan Science and Technology Program (Nos. 2019ZDZX0005, 2019ZDZX0006, 2020YFQ0056, 2021YFG0034), Talent Funding Project by the Organization Department of Sichuan Provincial Party Committee, and Science and Technology Service Network Initiative (KFJ-STS-QYZD-2021-21-001).

**Conflicts of Interest:** The authors declare that they have no conflicts of interest to report regarding the present study.

**Data availability statement:** The data that support the findings of this study are available on request from the corresponding author. The data are not publicly available due to privacy or ethical restrictions.